\documentclass[10pt]{article}
\usepackage[margin=1in]{geometry}
\usepackage{titling}
\usepackage{titlesec}

\usepackage{times}    
\usepackage[scaled=0.9]{helvet} 
\usepackage{sectsty}
\usepackage[T1]{fontenc}
\allsectionsfont{\sffamily}

\usepackage{graphicx}
\usepackage{todonotes}

\usepackage{subcaption}
\usepackage{url}
\usepackage{enumitem}

\usepackage[breaklinks]{hyperref}

\usepackage{ragged2e}
\justifying

\usepackage{xcolor}
\usepackage[numbers,square]{natbib}

\pretitle{
 \sffamily
 \vspace{-5em}
  \begin{center}
    \Large\bfseries
}
\posttitle{%
  \end{center}%
  \vskip .75em plus .25em minus .25em
}

\titleformat*{\section}{\large\bfseries}
\titleformat*{\subsection}{\normalsize\bfseries}
\titleformat*{\subsubsection}{\normalsize\bfseries}

\title{\Large\uppercase{Detailed Evaluation of Modern Machine Learning \\ Approaches for Optic Plastics Sorting}}

\author{\large 
    {\sffamily Vaishali Maheshkar - University at Buffalo, Dept of CSE} \\
    {\sffamily Aadarsh Anantha Ramakrishnan - NIT Trichy, Dept of CSE} \\
    {\sffamily Charuvahan Adhivarahan - University at Buffalo, Dept of CSE} \\
    {\sffamily Karthik Dantu - University at Buffalo, Dept of CSE}
}

\date{\vspace{-4ex}}

\begin{document}

\maketitle
\thispagestyle{empty}

\section{Abstract}\label{sec:abstract}

According to the \textcolor{black}{U.S. Environmental Protection Agency}, only a quarter of waste is recycled, and only 60\% of municipalities in the United States have curbside recycling pickup. Plastics have a \textcolor{black}{recycling} rate of only 8\%, \textcolor{black}{with an additional 16\% being incinerated}. \textcolor{black}{The other} plastics (76\%) are disposed of in landfills. Low plastic recycling rates are due to various factors such as contamination and \textcolor{black}{lack of incentive}, making it difficult to recycle them effectively. To boost efficiency, automated sorting is crucial. Companies like AMP Robotics and Greyparrot employ optical detection for automated sorting, while MRFs leverage Near IR sensors for precise plastic type detection~\cite{amp_plastics,greyparrot_plastics,titech_plastics}. 

Modern optical sorting, driven by advances in visual detection methods like object recognition and instance segmentation, incorporates modern Machine Learning algorithms. Two-stage detectors like R-CNN~\cite{girshick2014RCNN}, Fast R-CNN~\cite{girshick2015FastRCNN}, Faster R-CNN~\cite{ren2015FasterRCNN}, Mask R-CNN~\cite{he2017MaskRCNN} and SPP-net~\cite{yoshida2008SPPNet} employ region proposal and classification leveraging backbone architectures like ResNet~\cite{he2015deep}, CPSNet~\cite{wang2020CSPNet}, and EfficientNet~\cite{mingxing2019EfficientNet}. Single-stage detectors like YOLO\cite{redmon2016Yolo}, SSD~\cite{liu2016SSD}, and RetinaNet~\cite{lin2018focal} perform one-pass detection without a separate region proposal stage. While such methods excel under ideal conditions with a large volume of labeled training data, challenges arise in realistic scenarios, emphasizing the need to further examine the efficacy of optic detection for automated sorting. 

In this study, we have compiled several novel datasets to evaluate the efficacy of vision-based detection methods. These comprise over 20,000 images sourced from various sources. Through custom and public machine learning pipelines, this study aims to understand how optic recognition works as well as its limitations for sorting. Using metrics like Grad-CAM, saliency maps, and confusion matrices,  we highlight the features identified by optic sorting and their limitations. We perform this analysis on our custom trained model from the compiled datasets.

To conclude, our findings are that optic recognition methods have limited success in accurate sorting of real-world plastics at MRFs, primarily because they rely on physical properties such as color and shape. Further research is needed to adapt such models to identify plastics more accurately.  
\section{Introduction and Motivation}\label{sec:introduction}

Around the world, one million plastic bottles are purchased every minute, while up to five trillion plastic bags are used worldwide every year. Today, we produce about 400 million tonnes of plastic waste every year and the global production of primary plastic is estimated to reach 1,100 million tonnes by 2050~\cite{unep_plastics}. Approximately 36\% of all plastics produced are used in packaging, including single-use plastic products for food and beverage containers, approximately 85\% of which ends up in landfills or as unregulated waste~\cite{unep_plastics}.

The overall plastic recycling rate in the U.S. has historically been around 8-9\%, indicating that only a small fraction of the plastic generated is recycled. Different types of plastics have different recycling rates. For example, PET (polyethylene terephthalate) and HDPE (high-density polyethylene) containers, commonly used for beverage bottles and milk jugs, tend to have higher recycling rates compared to other types of plastics~\cite{epa_plastics}. Challenges in plastic recycling include the diversity of plastic types, contamination, and limited recycling infrastructure. Single-use plastics and complex multi-layer packaging often face lower recycling rates due to these challenges. 

Initiatives to enhance plastic recycling rates are in progress, incorporating advancements in sorting technologies like machine learning-based systems, heightened public awareness, and campaigns to minimize the usage of single-use plastics. In fact, one of the key objectives of the U.S. National Recycling Strategy~\cite{epa2021NationalRecyclingStrategy} from the EPA is to increase collection and improve materials management through the development of new research in recycling processes to make them more efficient to achieve a national goal of 50\% recycling rate by the year 2030. Optical Deep Learning-based sorting methods are of particular importance to us owing to their popularity, their pragmatic benefits like real-time applicability and tolerance to input variations and most importantly their impressive accuracy results. \textcolor{black}{These systems, deployed by companies like AMP Robotics, Greyparrot, Tomra Sorting Solutions~\cite{tomra}, utilize sophisticated cameras, sensors, and machine learning algorithms to identify and sort plastics in recycling streams. They analyze visual cues such as color, shape, and texture to differentiate between different types of plastics, enabling high-speed and accurate sorting processes. However, alongside optical recognition, emerging alternative technologies such as digital watermarking are being tested at scale. Companies like P\&G and BASF have begun exploring the use of digital watermarks~\cite{pg}~\cite{basf} embedded in plastic products to facilitate identification and sorting during the recycling process. By leveraging optical scanners and specialized detection algorithms, recycling facilities can extract and interpret these watermarks to streamline sorting operations.}

\textcolor{black}{Through a critical examination of these commercially available technologies, including optical recognition systems and emerging alternatives like digital watermarking, this study seeks to discern their strengths, limitations, and potential synergies in advancing plastic recycling efforts.} With any black box method, it is not only important to understand how well these methods work but also how they work to understand the trade-offs they make. This study delves into an analysis of the specific regions prioritized by contemporary machine learning algorithms in the plastic classification process. The findings lead us to propose the adoption of a multi-modal system to further enhance the efficiency and accuracy of plastic sorting.

\section{Review of Related Work}\label{sec:related_work}

The review of related work reveals a multifaceted landscape in the domain of plastic waste classification and recycling. Various approaches have been employed to tackle the challenges associated with identifying and segregating plastics based on their physical properties. One notable method involves utilizing visual and physical properties to classify plastics, leveraging features such as weight, pressure, and color. \textcolor{black}{Machine learning algorithms, including SVM, KNN, Decision tree, and Logistic Regression, have been employed to analyze these features and classify plastics into recycled or non-recycled categories}~\cite{kambam2019Classification}. 

Another technique using physical features of plastics involves adapting image sensors and deep learning object detection algorithms, such as the YOLO~\cite{redmon2016Yolo} model, to enhance plastic waste classification based on the shape of the waste. This approach aims to improve accuracy, especially when dealing with materials having similar chemical compositions but differing physical attributes. \textcolor{black}{In experiments, they achieved a classification accuracy rate exceeding 91.7\% mean Average Precision (mAP) in distinguishing between PET and PET-G} 
~\cite{choi2023PlasticWasteClassification}. Additionally, ZeroWaste~\cite{bashkirova2022Zerowaste}, an in-the-wild industrial-grade waste detection and segmentation dataset, describes the formation of a real world dataset, addressing the challenges of the complex and cluttered nature of waste streams. 

Furthermore, the exploration of one-shot learning techniques in image-based classification of plastic waste, achieving an accuracy of 99.74\%, showcases the solutions in identifying and segregating plastics for recycling using Machine Learning~\cite{agarwal2009OneShotPlastic}. \textcolor{black}{Multi modal approaches using Deep Learning and plastic spectral information have also been explored for plastic sorting. In ~\cite{neo2023cross}, the authors introduce Multi-modal Plastic Spectral Database (MMPSD) utilizing chemometric analysis of plastic spectral data coupled with deep learning techniques. The MMPSD provides a comprehensive repository of Fourier Transform Infrared (FTIR), Raman, and Laser-induced Breakdown Spectroscopy (LIBS) data for each sample in the database. Using Spectral Conversion Autoencoders (SCAE) technique, they achieved an increase in the classification accuracy as compared to an uni-modal approach from 0.933 to 0.970.}

These diverse approaches collectively contribute to the ongoing efforts to improve recycling efficiency and sustainability in waste management systems. \textcolor{black}{Overall, while all the approaches aim to improve plastic waste classification and recycling efficiency, they differ in the techniques used, the properties considered for classification, and the technologies employed for data capture and analysis.} However, through our analysis of the highly performant optical deep learning methods in this work, we wish to show that the visual features that they pay attention to need not necessarily be direct indicators of plastic type.

\section{Technology Approach}\label{sec:approach}

\textbf{}
In our approach, we claim that most image-based plastic classification algorithms are based on identifying the physical characteristics of plastics. We also use various analysis tools such as Grad-CAM ~\cite{Selvaraju_2019}, Mean Average Precision (mAP) and Confusion Matrices to prove our claim. To evaluate the efficacy of various vision-based detection algorithms, we have aggregated images of plastics from various sources.

\begin{figure}
    \centering
    \includegraphics[trim={0 20cm 2cm 1cm},clip,width=\textwidth]{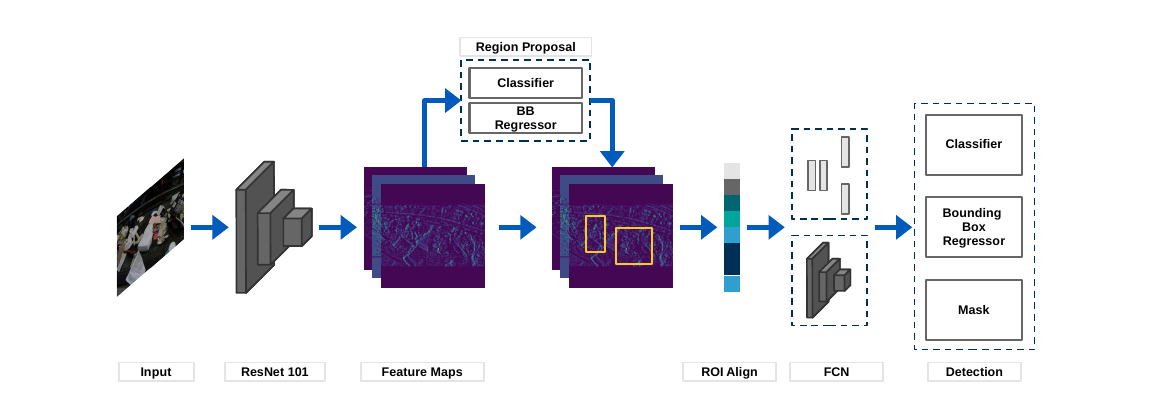}
    \caption{The architecture of Mask RCNN~\cite{he2017MaskRCNN} with its components, a popular Deep Learning architecture used to identify and classify multiple objects of interest on color images.}\label{fig:maskrcnn-architecture}
\end{figure}

\begin{enumerate}
    \item \textbf{Web-scraped Plastics Dataset:} We have collected a dataset of 9440 images across 7 plastic types (HDPE, LDPE, PVC, PET, PS, PP and Other) procured through web scraping from online sources like Google, Bing and Yandex (labeled). We applied a pretrained Resnet-34 CNN model and achieved a classification accuracy of 96\%. The images obtained through web scraping are images of clean plastic samples without any deformations, which unfortunately are not representative of the real world.
    \item \textbf{United States Plastic Corp. Dataset:} We also collected images from the United States Plastic Corp. catalogs to further increase our image dataset consisting of different plastic types and were able to extract 9564 images of 6 types of plastics (HDPE, LDPE, PET, PP, PS, PVC) through web-scraping. For this vast dataset we have collected, we applied the Mask RCNN algorithm to identify the objects in the images and achieved a mean Average Precision (mAP) of around 80\% for the validation dataset. This dataset is also clean (fresh from the factory) which made us look further, into collecting a dataset having real-world plastic samples with impurities and deformations.
    \item \textbf{Open Food Facts Dataset:} We came across Open Food Facts, a non-profit organization working towards creating an extensive food product database collected using their app. This dataset has real-world samples, and all the images are open source, sent in by volunteers around the world. These images have been collected in challenging environments (different backgrounds, lighting conditions etc.) and we used their APIs to get images, which have been validated using their corresponding plastic resin code logos. 
    
    We collected around 2721 images of different plastic types and after applying the pretrained Resnet-34 model with transfer learning, the accuracy obtained is 71.8\%. The images in this dataset mainly focus on the food product label with the food specifications, but the images have been manually cleaned to ensure that the food packaging is also visible. Also, the dataset is limited to food products and does not contain any deformed plastics.
    \item \textbf{Material Recycling Facility dataset:} In order to collect real-world data of deformed and contaminated plastics for efficient plastics identification and classification using computer vision, we visited the local Material Recycling Facility in Buffalo and set up cameras on multiple conveyor belts for data collection. We collected data from two different conveyor belts, one carrying only PET samples and the other carrying mixed plastic samples. We used GoPro Hero11 Black Camera at 2.7K240 and 5.3K60 resolution and collected around 800 images from the moving conveyor belts with the plastic objects.
\end{enumerate}

\subsection{\textbf{Machine Learning on the Datasets}}

Below are some of our findings about the computer vision algorithms for plastics sorting - empirical evidence suggests that they rely on properties such as shape, color and object labels for classification of plastics. We use various metrics such as Grad CAM activations, mAP and confusion matrix to highlight the features used by such methods for classification. 

\subsubsection{Application of Different Machine Learning Algorithms to plastic image dataset}

\begin{enumerate}
    \item \textbf{Mask RCNN on the MRF Dataset}
    
    Mask RCNN (Mask Region-based Convolutional Neural Network) is a popular deep learning model used for instance segmentation tasks, which identifies and delineates objects at the pixel level within an image. It is an extension of the Faster R-CNN architecture and is known for its ability to simultaneously predict object bounding boxes, class labels, and pixel-level masks for each instance in an image [20]. \textcolor{black}{Figure 1 shows the architecture of Mask RCNN with its various components.} We trained 70 (50 train and 20 validation) images of the MRF dataset using Mask RCNN algorithm to identify the different plastic objects in the image and get the activation maps to identify which regions of the image the algorithm is focusing on.

\begin{figure}
    \centering    \includegraphics[width=0.56\textwidth]{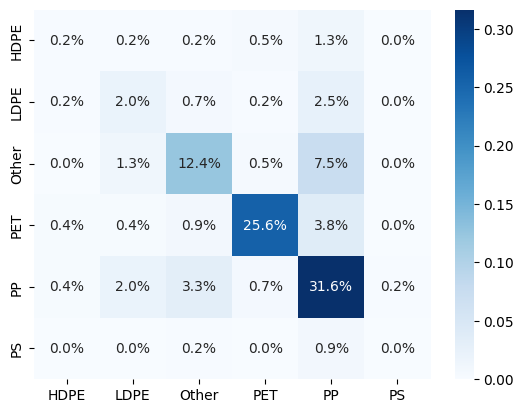}
    \caption{Confusion matrix for the OFF dataset. Higher numbers along the diagonal indicate proper classification. Off-diagonal elements indicate misclassification. For example, we can observe that a significant percentage of PP plastic is being classified as LDPE, PET or other}
    \label{fig:confusion-matrix}
\end{figure}
    
    \item \textbf{Mask RCNN on the US Plastics Dataset}
    
    \textcolor{black}{We also applied Mask RCNN on the US Plastics Corp. dataset to evaluate performance across different datasets, which allows for direct comparison of model effectiveness and generalization capabilities.} We trained 6690 images and validated 2874 images for classification and achieved a mean average precision (mAP) of around 80\% for the validation dataset. The mAP indicates the good performance of the Mask RCNN algorithm for object detection on the dataset images. \textcolor{black}{We evaluated Mask RCNN on the US Plastics dataset to compare the activations across different datasets.} 
    
    \item \textbf{Resnet-34 on the Open Food Facts Dataset}
    
    We obtained the below metrics for training and testing on the Open Food Facts images. Accuracy: 71.8\% Precision: 71.6\%, Recall: 71.8\%, F1 score: 71.1\%. \textcolor{black}{Figure 2 shows the confusion matrix for the OFF dataset. From the confusion matrix we see that there are very low correct predictions in PS and HDPE, due to class imbalance which is commonly observed in real-world datasets. ResNet-34, a variant of the Residual Neural Network (ResNet) architecture, was selected as the baseline model for comparison purposes to evaluate the performance and behavior of more complex models such as Mask RCNN in relation to a simpler architecture.  By employing ResNet-34 as the baseline model, we aim to understand what the Convolutional Neural Network (CNN) learns during the training process. The relatively lightweight nature of ResNet-34 allows for easier analysis of learned features and representations within the network and adds to the intuition with other results in this section.}

\begin{figure}
    \centering
    \begin{subfigure}{0.45\textwidth}
        \includegraphics[width=\linewidth]{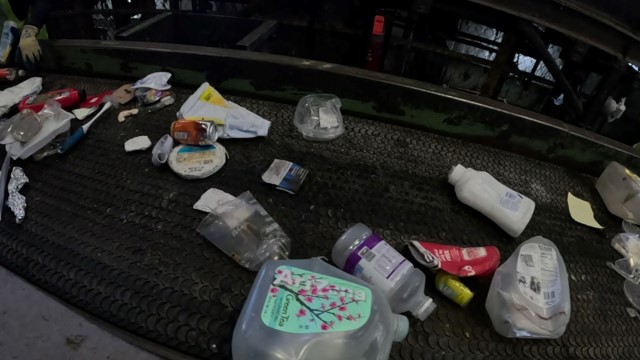}
        \caption{ }
        \label{fig:mrf_image3}
    \end{subfigure}
    \hfill
    \begin{subfigure}{0.45\textwidth}
        \includegraphics[width=\linewidth]{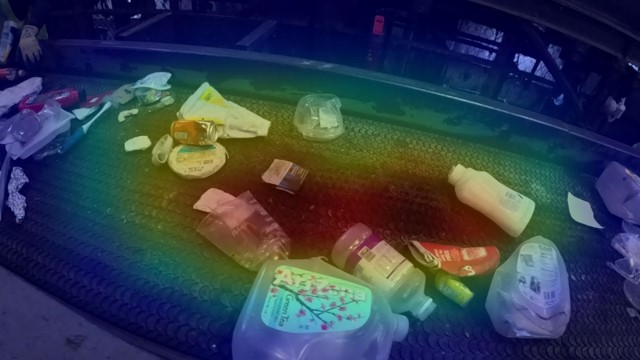}
               \caption{ }
        \label{fig:mrf_activations2}
    \end{subfigure}
    \\
    \begin{subfigure}{0.45\textwidth}
        \includegraphics[width=\linewidth]{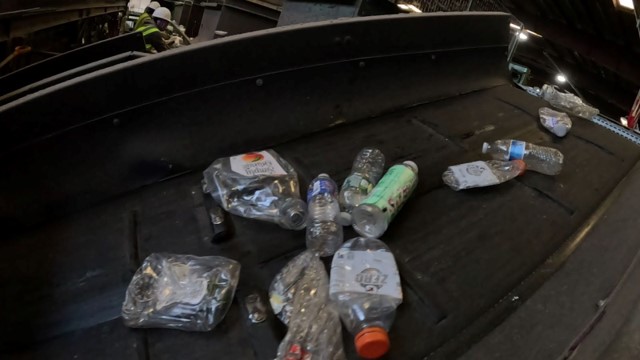}
               \caption{ }
        \label{fig:mrf_image}
    \end{subfigure}
    \hfill
    \begin{subfigure}{0.45\textwidth}
        \includegraphics[width=\linewidth]{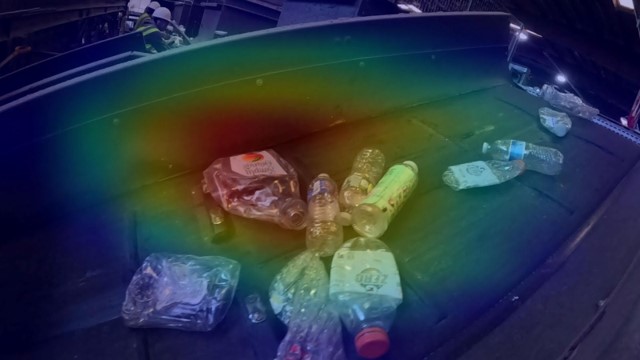}
               \caption{ }
        \label{fig:mrf_activations3}
    \end{subfigure}
    \\
    \begin{subfigure}{0.45\textwidth}
        \includegraphics[width=\linewidth]{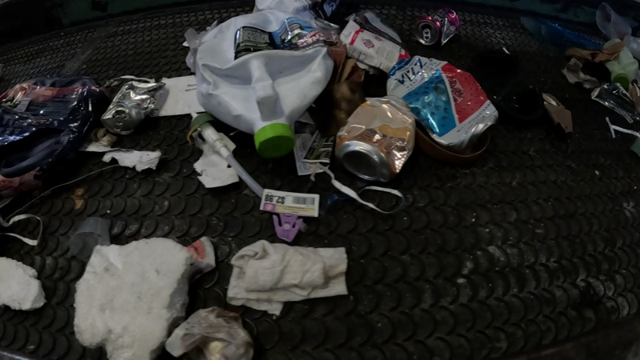}
               \caption{ }
        \label{fig:mrf_image5}
    \end{subfigure}
    \hfill
    \begin{subfigure}{0.45\textwidth}
        \includegraphics[width=\linewidth]{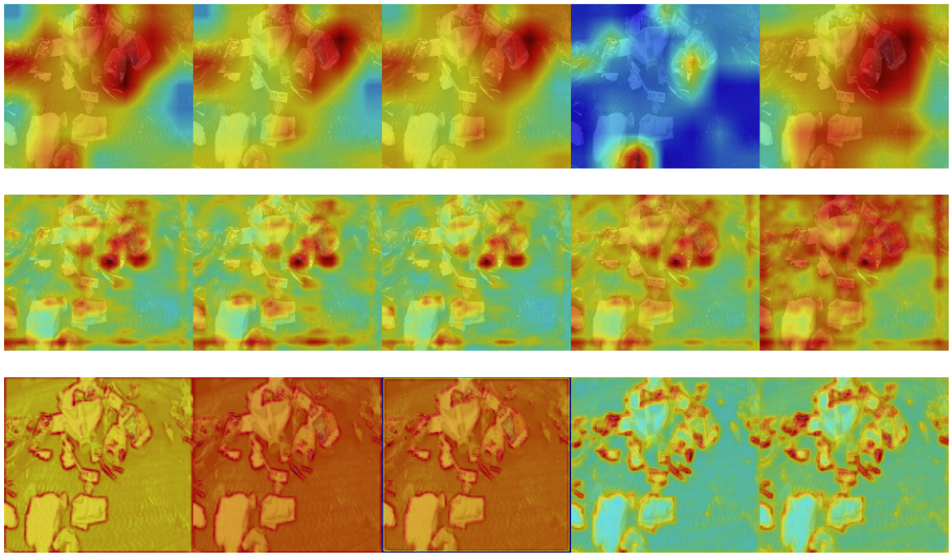}
        \caption{ }
        \label{fig:mrf_heatmaps}
    \end{subfigure}
    \caption{Grad CAM activations for samples from the MRF dataset, the activations show the regions highly focused by the network for object detection, in this image it localizes the objects, but the dark red region shows the highlights are on conveyor belt which should not play a role in classification. Heatmaps (f) for sample (e) show that focus is on the metal cans for most of the layers, likely due to optical feature variations.}
    \label{fig:mrf_comparison}
\end{figure}
    
    \item \textbf{Xception Model on the MRF Dataset}
    
    \textcolor{black}{Xception model, a deep learning convolutional neural network architecture designed to enhance the computational efficiency by using depthwise separable convolutions and pre-trained on the ImageNet dataset, was used to get activations on the MRF dataset. It aims to capture intricate patterns while reducing the number of parameters and is used for various computer vision tasks, including image classification, object detection, and semantic segmentation. This model was used to evaluate the Grad CAM activations and compare it with the activations from the Mask RCNN model. From the results in Figure 3 (a,b,c,d) we see that the focus of the model is on the conveyor belt region.} 
    
    \item \textbf{Resnet-50 on the MRF Dataset}
    
    Resnet-50, a variant of the ResNet (Residual Network) architecture, designed to address the challenges of training very deep neural networks, was pre-trained on the ImageNet dataset and used to get activations on the MRF dataset. \textcolor{black}{This architecture allows us to capture different aspects of the MRF dataset, as there are variations in model design and complexity. It also provides insights into the robustness of the model by providing consistent predictions as the other models on the MRF dataset. The MRF dataset being a challenging one with deformed and contaminated plastic images, enables a comprehensive exploration of the dataset, facilitates performance comparison, aids in model selection, and enhances insights into model behavior. From the Grad CAM activations in Figure 3 (e,f) for this model, we see that the activations are focused on the conveyor belt instead of the plastic samples.}
\end{enumerate}

\begin{figure}
    \centering
    \includegraphics[width=0.6\textwidth]{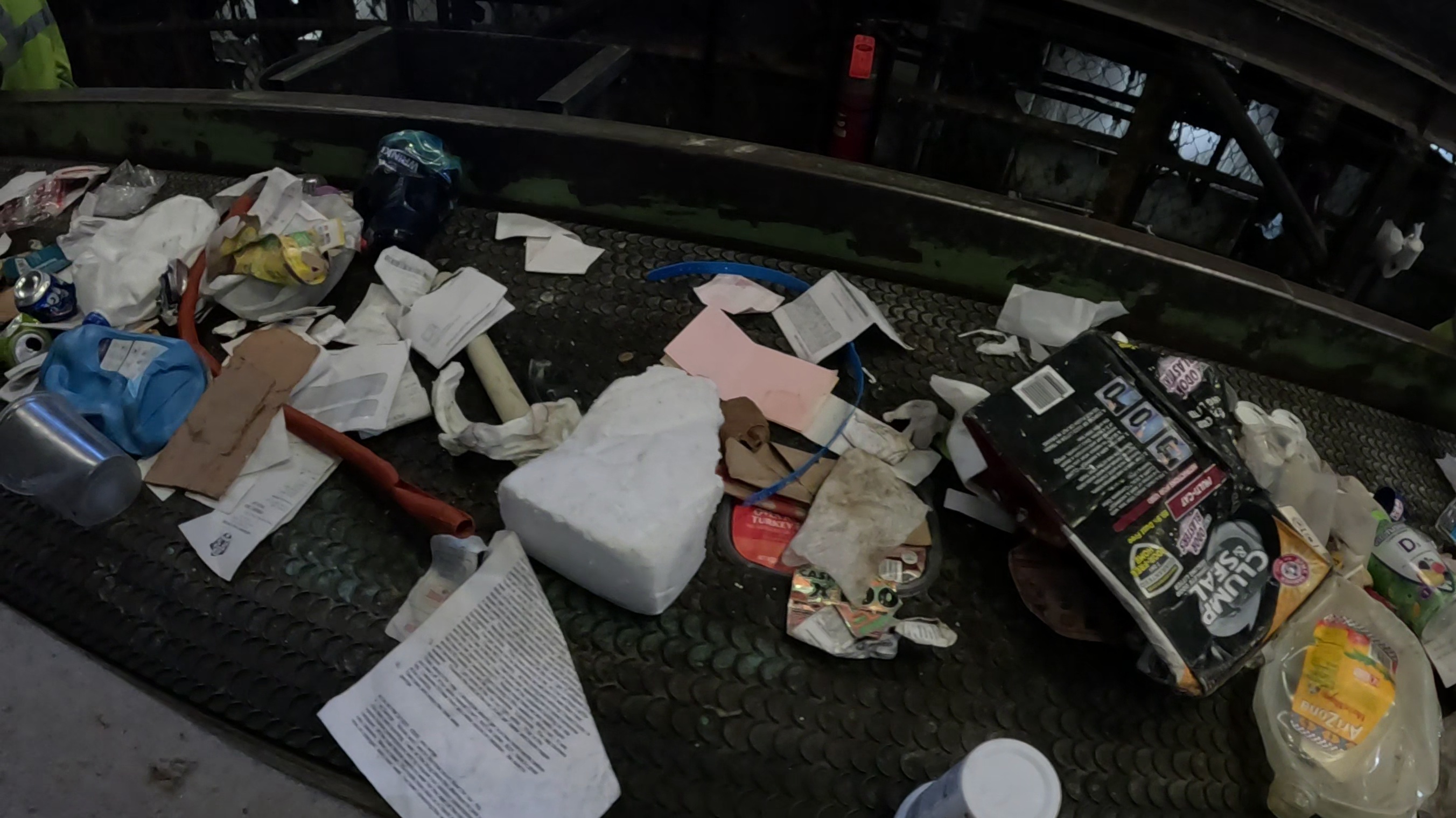}
    \caption{A sample image collected from a regional Materials Recovery Facility (MRF). Materials are typically mangled, mixed, sometimes occluded and are transported through a fast conveyor system where they are sorted with automated and human-in-the-loop processes. MRF dataset collection posed challenges, with camera placement away from the conveyor belt to reduce vibrations and height limitations.}\label{fig:mrf-sample-1}
\end{figure}
\vspace{0cm}
\begin{figure}
    \centering
    \includegraphics[width=0.8\textwidth]{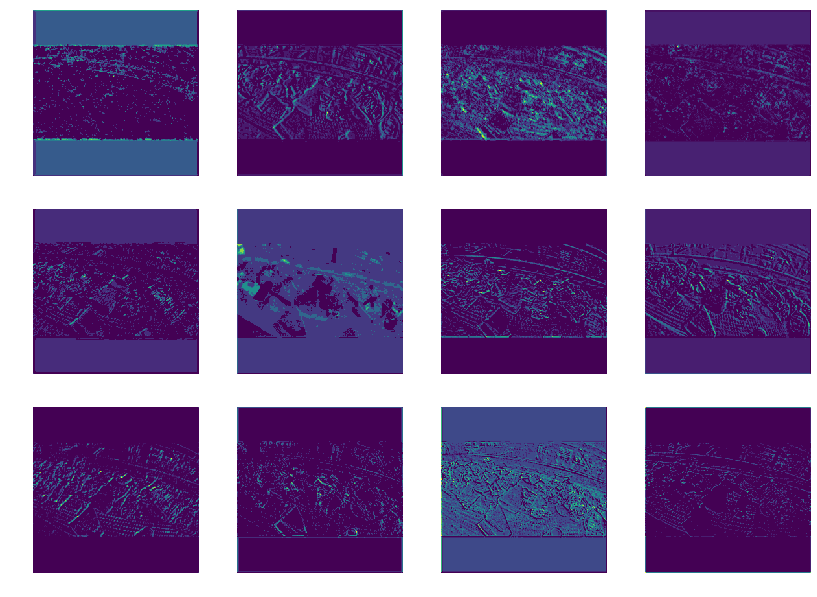}
    \caption{Feature maps obtained from running the image sample from~\autoref{fig:mrf-sample-1} through the Mask RCNN model. The brighter regions in these feature maps indicate higher weighting in comparison to darker regions. One can interpret that these regions were used to identify plastics as a certain type and be used to distinguish from other types.}
    \label{fig:mrf-maskrncc}
\end{figure}

\vspace{-1em}
\subsubsection{Analysis Method}

\begin{enumerate}
    \item \textbf{Feature Maps} in deep learning and neural networks represent the output of intermediate layers in the network. These maps capture the presence of specific features or patterns within the input data. Each feature map corresponds to a particular filter or kernel applied to the input data.
    
    \item \textbf{Grad CAM} (Gradient weighted Class Activation mapping) is used to visualize the regions of an image that contribute the most to the prediction made by a convolutional neural network (CNN), it provides insights into what parts of the input image the model is focusing on when making a particular prediction.
\end{enumerate}

\textcolor{black}{Figure 3 (a,b,c,d) shows the Grad CAM activations from the Xception model, and we see that the focus is on the conveyor belt region. Furthermore, for Figure 3 (e,f) which shows the Grad CAM activations from the Resnet50 model, we see that the activations are in the conveyor belt region and towards the metal cans as against the HDPE milk can adjacent to the metal cans. Figure 4 is a sample image used for getting activations. Figure 5 represents the feature maps obtained by applying Mask RCNN on the sample image. Figure 6 represents a repetition of the above process used, for another sample image. From the feature maps shown, we can infer that the Mask RCNN focuses on the conveyor belt for most of the intermediate layers, instead of the plastic material itself.
Similarly, for the US Plastics dataset, Figure 7 represents the feature maps that focus on the shape of the object.
Figure 8 shows the Grad CAM activations for the OFF dataset, we see that for most of the activations, the focus is on the label of the plastics.}

\begin{figure}
    \centering
    \includegraphics[width=0.5\textwidth]{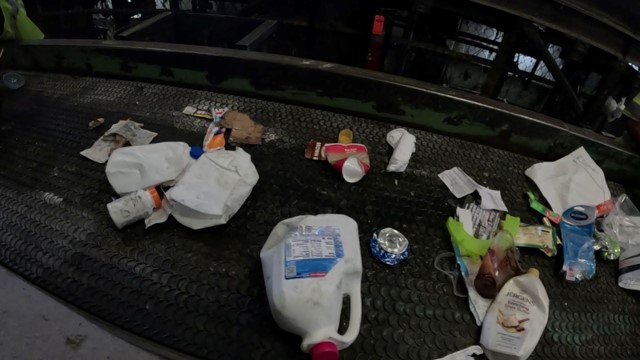}
    \includegraphics[width=.28\textwidth]{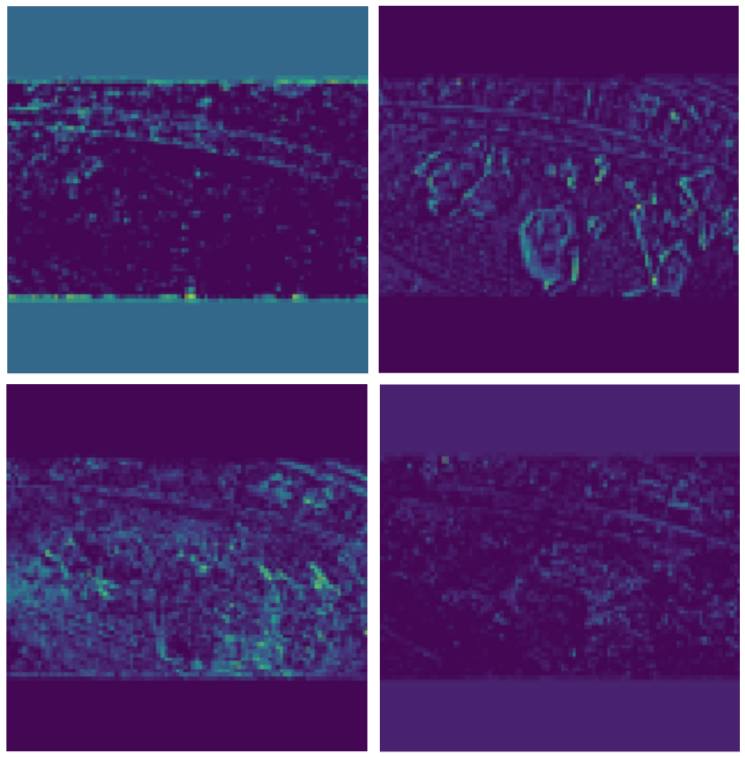}
    \caption{Another sample image from the MRF mixed plastics conveyor belt (left). Real-world images are challenging with high variations in the number of items, their positions, size, orientation, occlusion, and impurities etc. Feature maps (right) indicate that convolutional models are adept at handling most of these variations.}\label{fig:mrf-sample-2}
\end{figure}

\vspace{0cm}

\begin{figure}[!htb]
    \centering
    \begin{subfigure}{0.2\textwidth}
    \begin{subfigure}{\textwidth}
        \includegraphics[width=\linewidth]{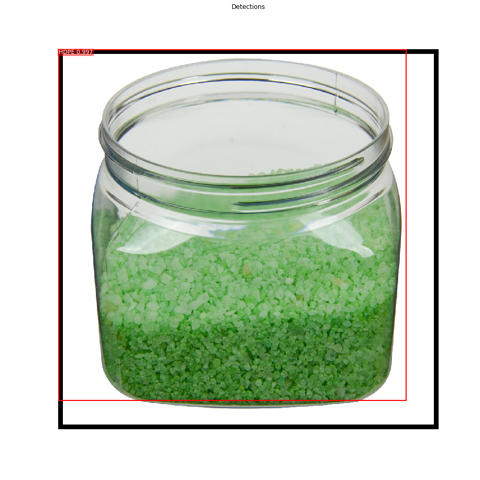}
        \caption{ }
    \end{subfigure}
    \\
    \begin{subfigure}{0.9\textwidth}
        \includegraphics[width=\linewidth]{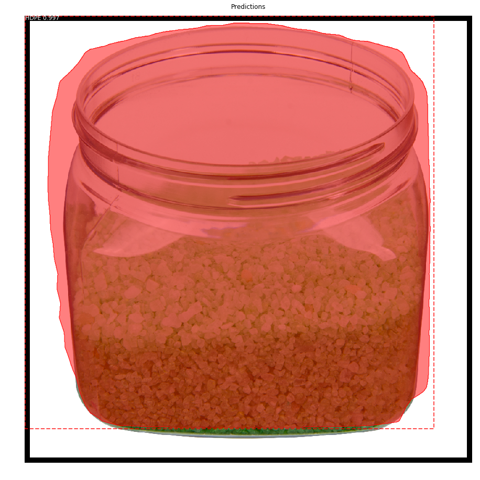}
        \caption{ }
    \end{subfigure}
    \end{subfigure}
    \begin{subfigure}{0.6\textwidth}
    \includegraphics[width=\textwidth]{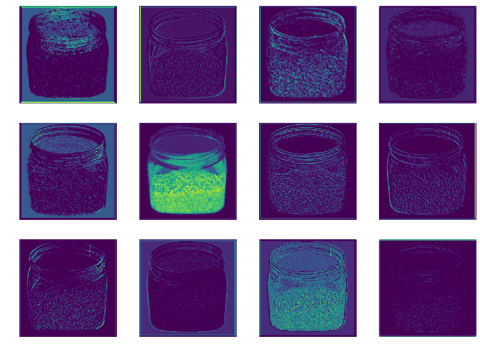}
    \caption{ }
    \end{subfigure}
    \caption{What do optical DNN models prioritize to determine the plastic class? (a) Object Detection using Mask RCNN (b) Model Prediction using Mask RCNN (c) Feature maps, i.e. the portions of images that contributed highly towards the classification result, indicate that shape, color and sometimes contents are used. }\label{fig:activations}
\end{figure}

\begin{figure}
    \centering
    
    \begin{minipage}{0.23\textwidth}
        \includegraphics[width=\linewidth, height=4cm]{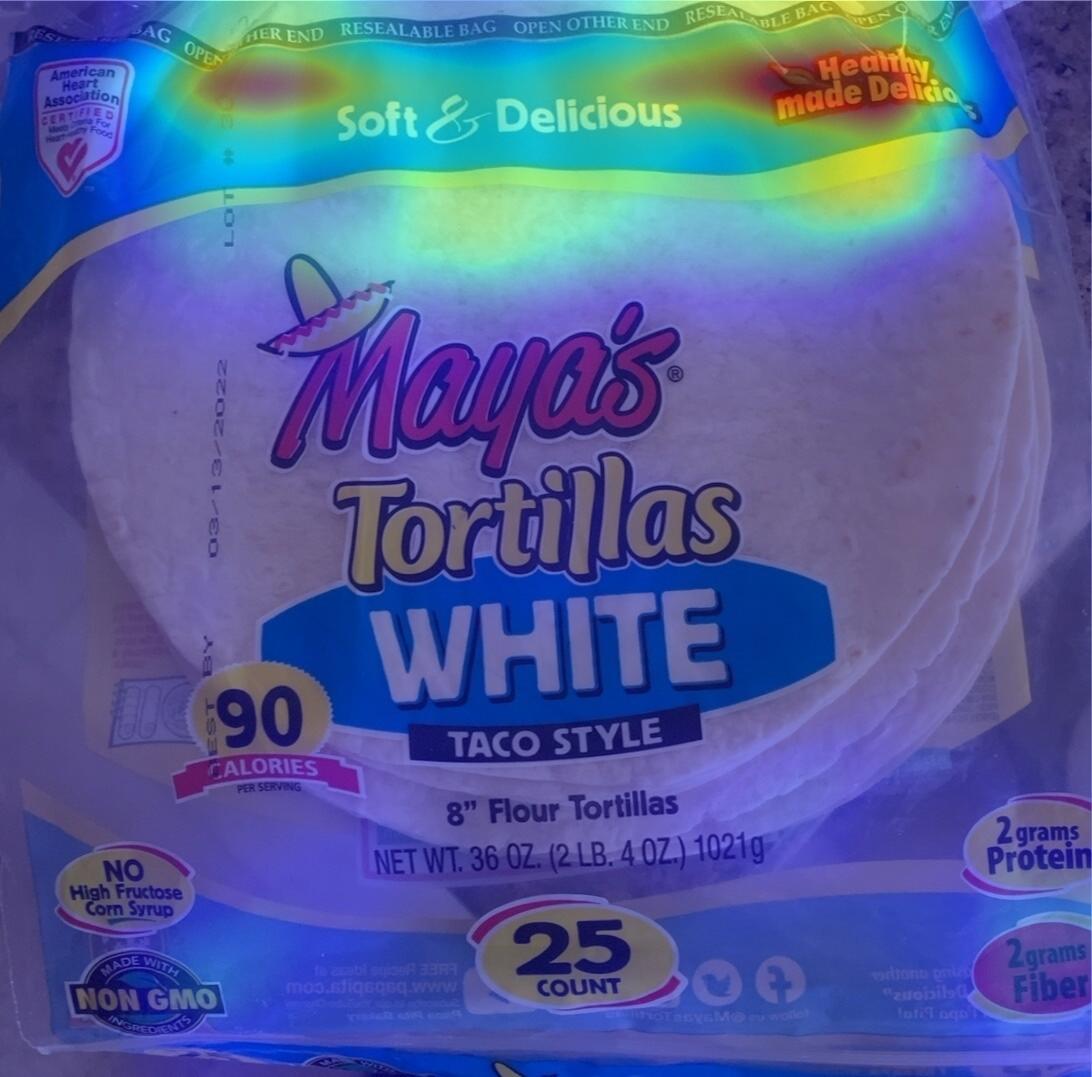}
    \end{minipage}
    \begin{minipage}{0.23\textwidth}
        \includegraphics[width=\linewidth, height=4cm]{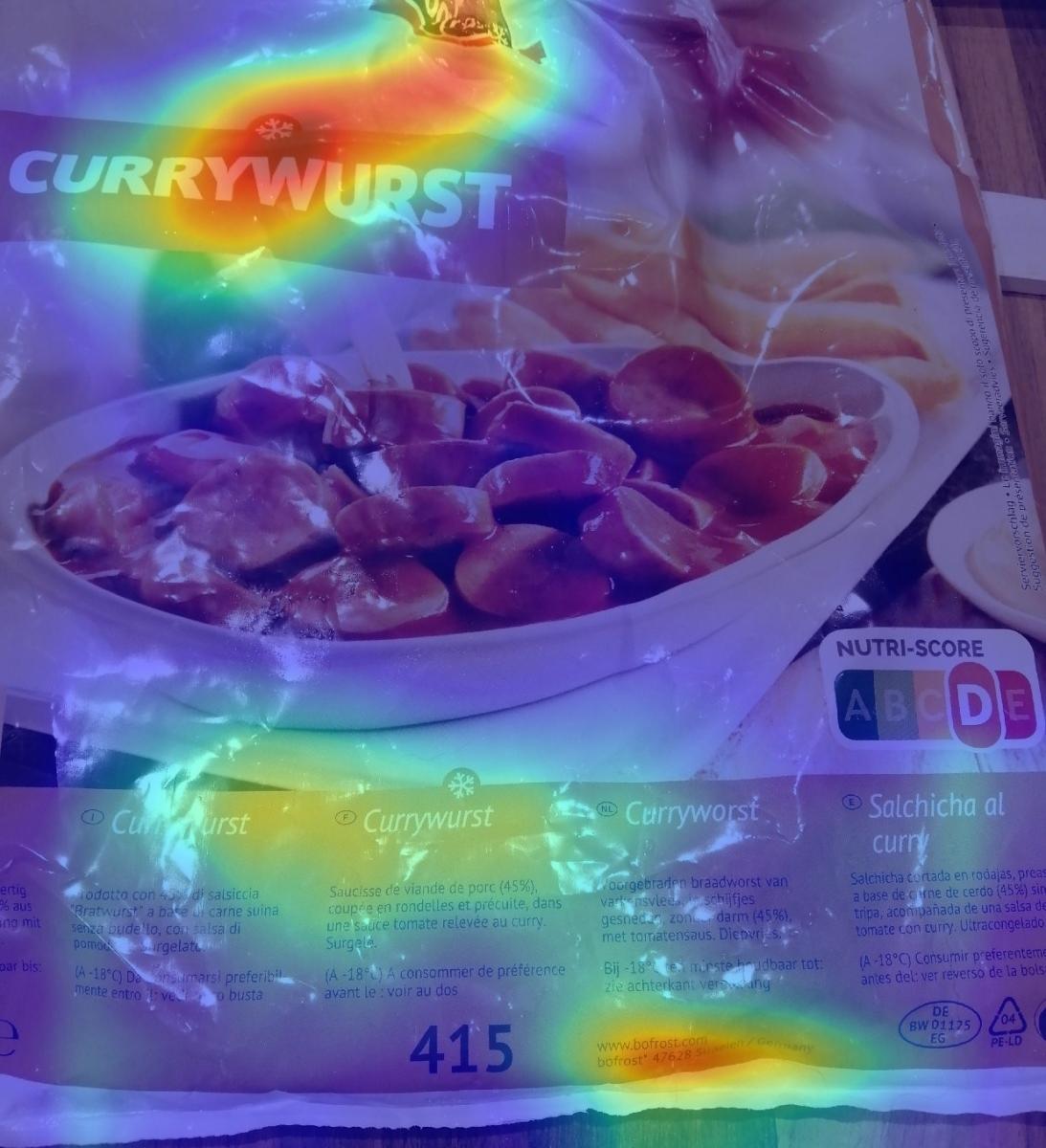}
    \end{minipage}
    \begin{minipage}{0.23\textwidth}
        \includegraphics[width=\linewidth, height=4cm]{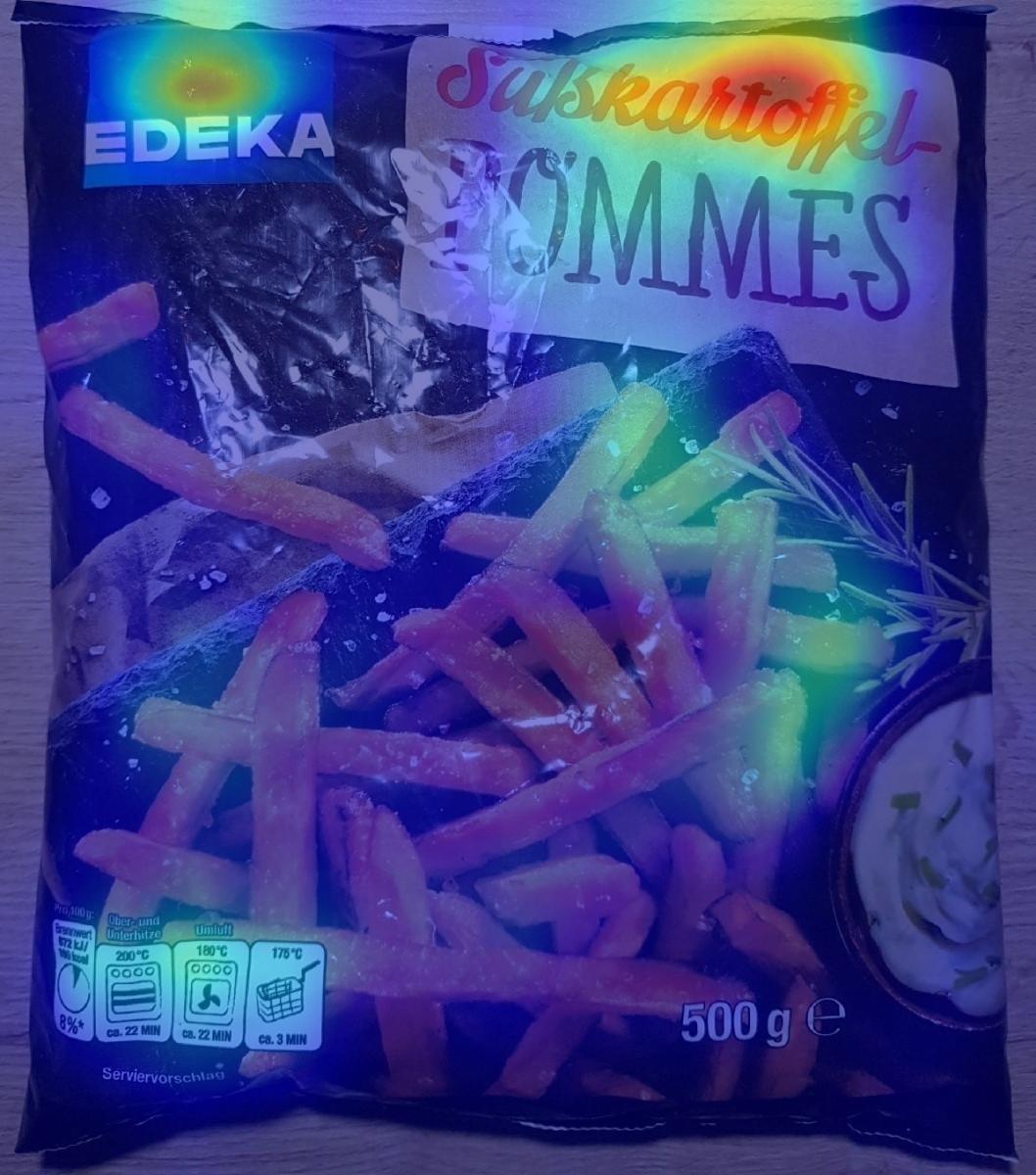}
    \end{minipage}
    \begin{minipage}{0.23\textwidth}
        \includegraphics[width=\linewidth, height=4cm]{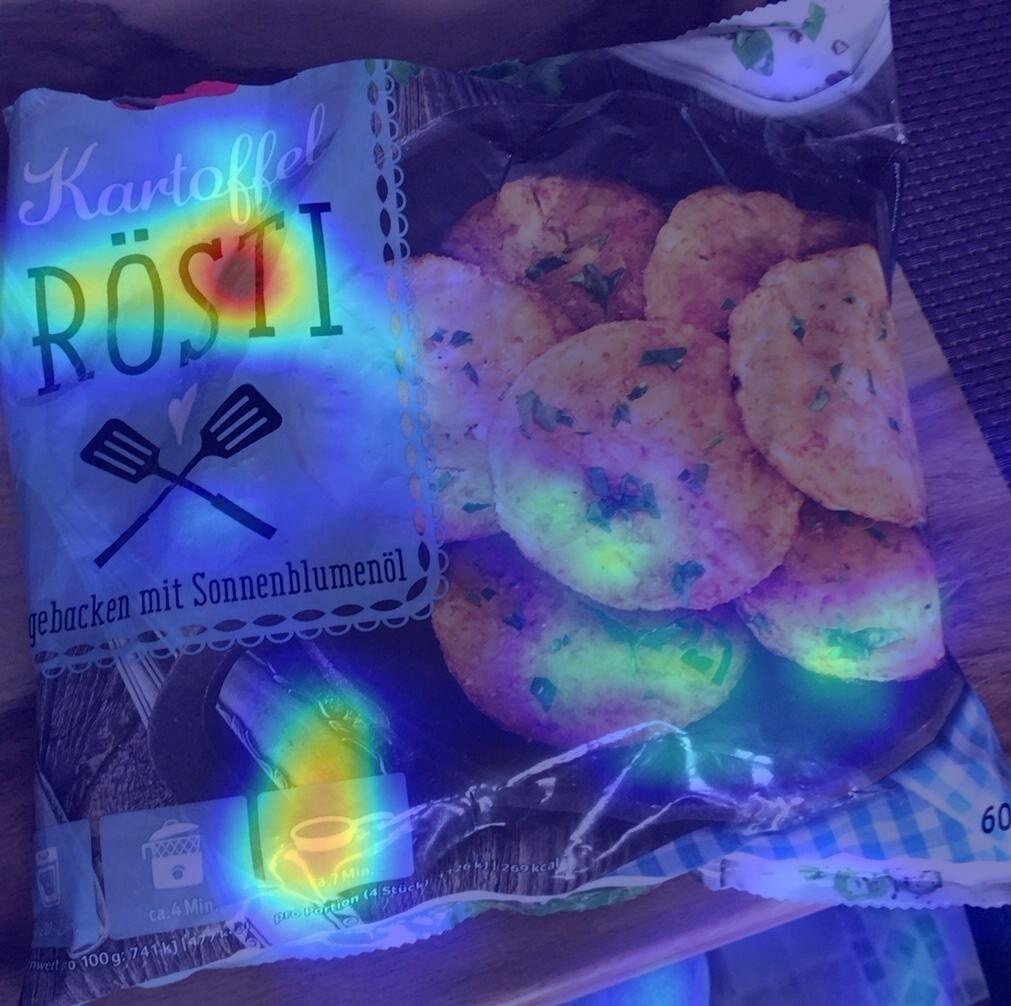}
    \end{minipage}
    \\
    \begin{minipage}{0.23\textwidth}
        \includegraphics[width=\linewidth, height=4cm]{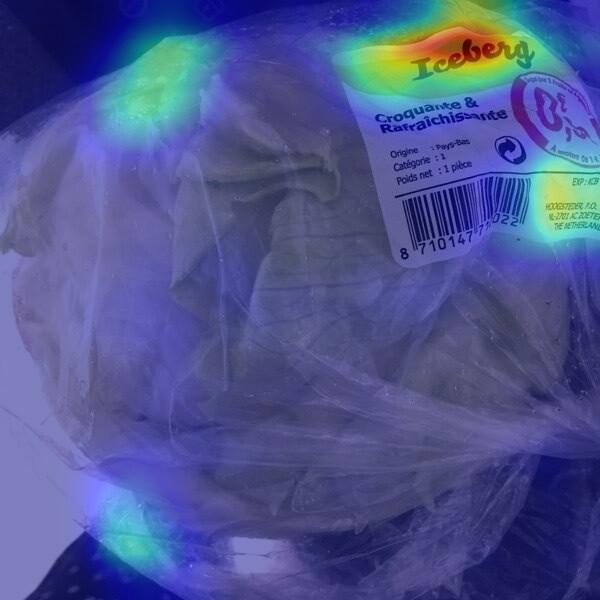}
    \end{minipage}
    \begin{minipage}{0.23\textwidth}
        \includegraphics[width=\linewidth, height=4cm]{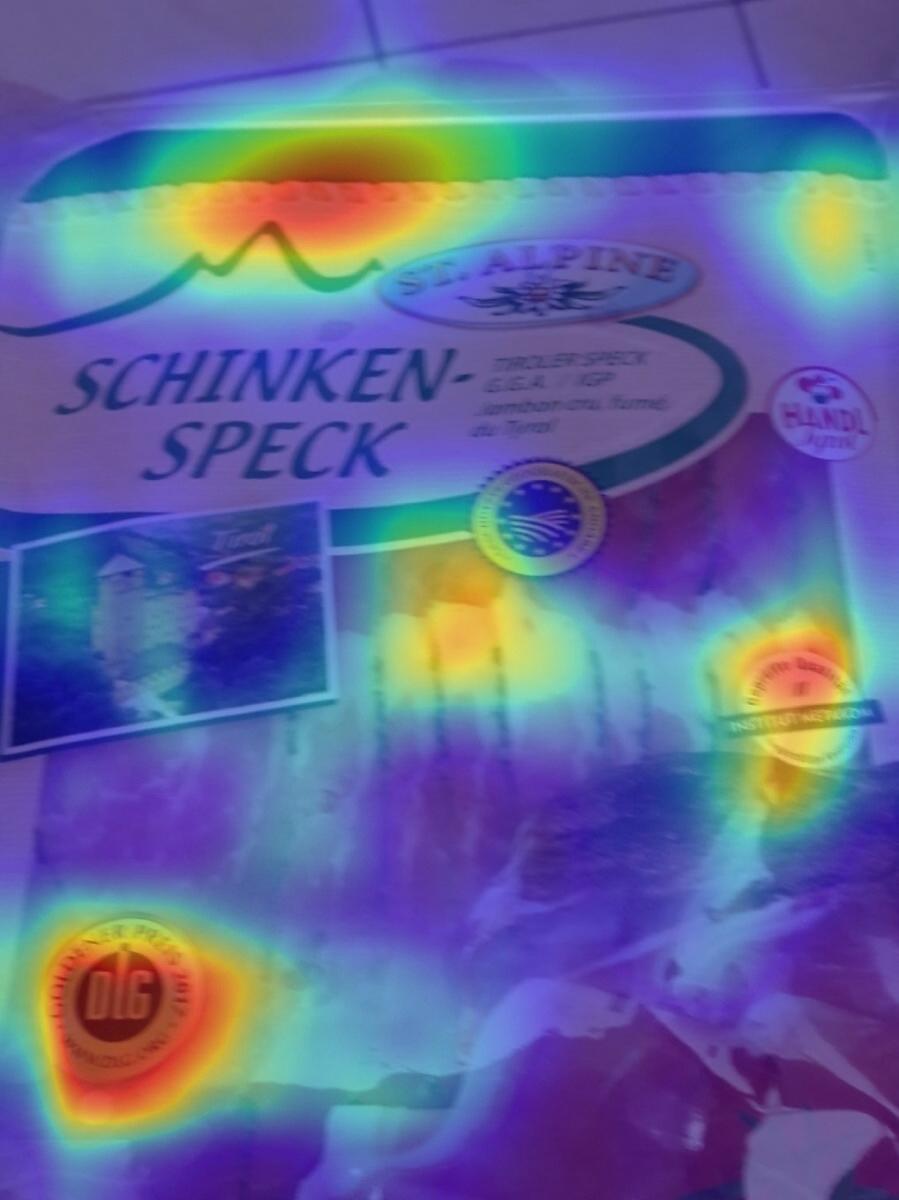}
    \end{minipage}
    \begin{minipage}{0.23\textwidth}
        \includegraphics[width=\linewidth, height=4cm]{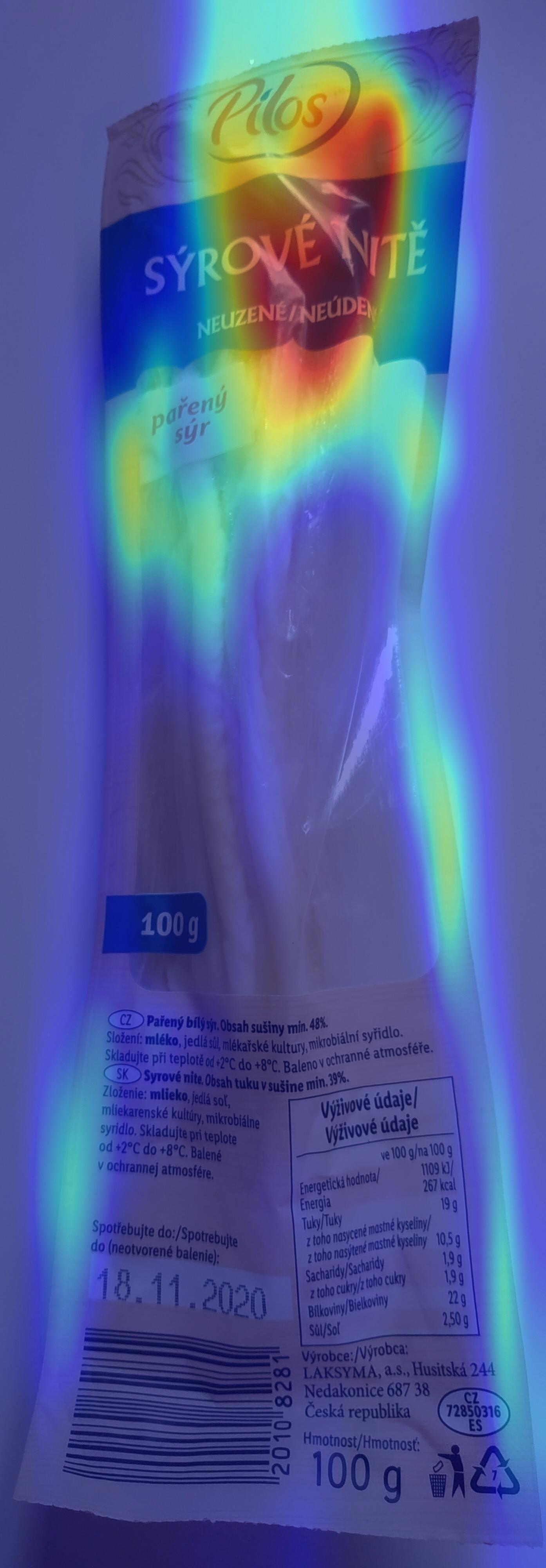}
    \end{minipage}
    \begin{minipage}{0.23\textwidth}
        \includegraphics[width=\linewidth, height=4cm]{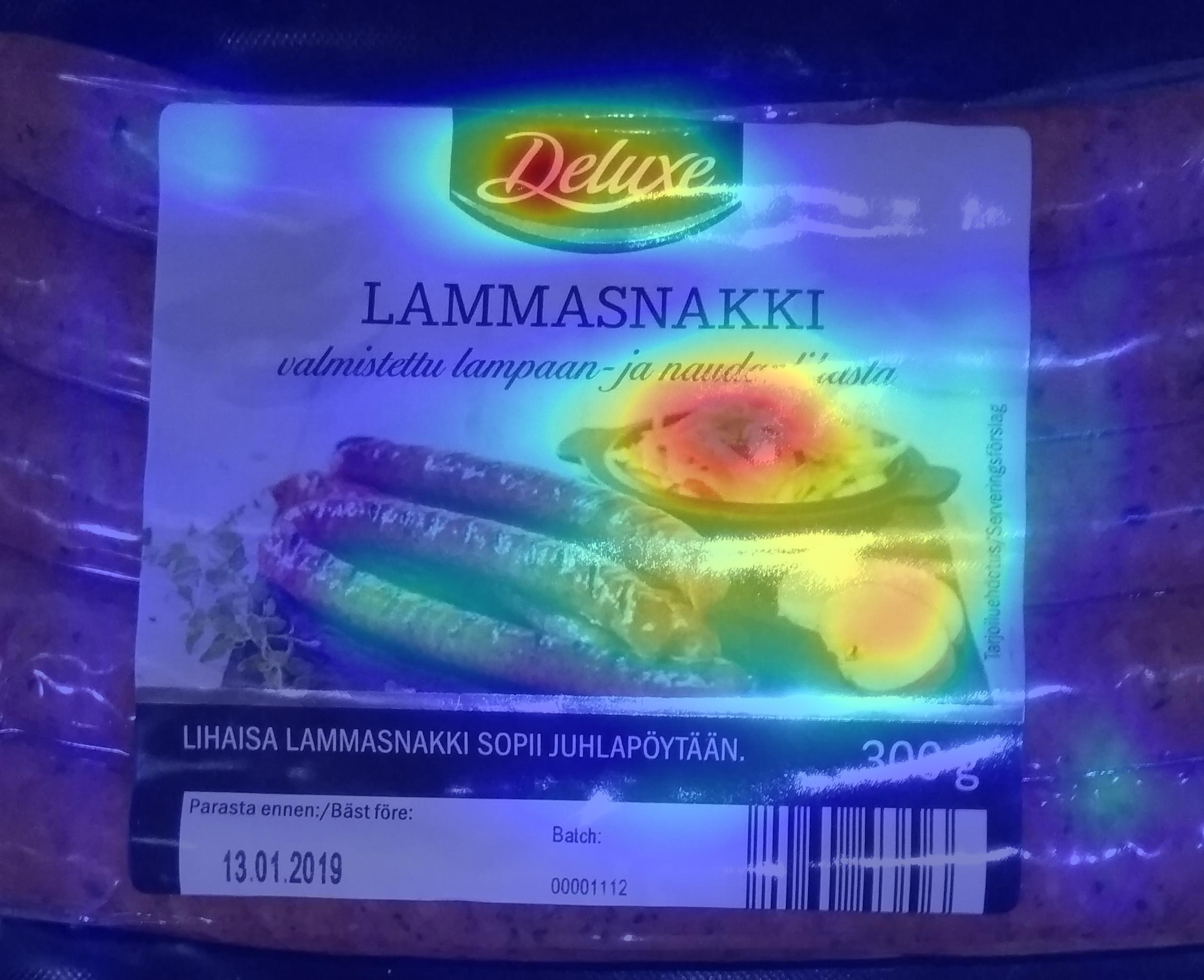}
    \end{minipage}
    
    \caption{Grad CAM activations for OFF dataset using Resnet-34. The red regions in the image contributed highly towards the classification. We observe that many of these regions are on visual patterns on the packaging like logos, images and sometimes contents of the packaging for transparent plastics. The regional shape, color and contrast variations form prime contenders for discriminative models.}
    \vspace{-1em}
\end{figure}

\section{Discussion}\label{sec:discussion}

The observations from our machine learning experiments on plastic sorting using computer vision algorithms reveal intriguing insights into the limitations and strengths of the applied models. In the case of Mask RCNN trained on the MRF dataset, despite successfully identifying plastic objects, the activation maps demonstrate a predominant focus on the conveyor belt rather than the plastic type across various layers. \textcolor{black}{Addressing black colors in images required manual intervention, adding complexity to deep learning model processing.} This emphasizes strongly about the models accurately discerning plastic materials based on shape and color. Also, when Mask RCNN is trained on the US Plastics dataset, the model exhibits commendable accuracy in predicting plastic types, with a significant focus on the shape of the objects. This suggests that the model effectively learns discriminative features for classification, aligning with its high mean average precision (mAP) of around 80\%. On similar lines, Resnet-34 trained on the Open Food Facts dataset, while correctly predicting plastic types, the Grad CAM activations indicate a focus on the labels of plastic objects. Finally, Grad CAM visualizations obtained using the Xception model pre-trained on the Imagenet dataset reveal a noteworthy pattern — the model concentrates on the conveyor belt rather than the objects for classification. These observations underscore the complexity of accurate plastic sorting, urging further investigation into refining models to enhance their reliability in real-world scenarios. 
\section{Conclusion and Recommendations}\label{sec:conclusion}

In this study, our contribution lies in curating a comprehensive image dataset, comprising both labeled and unlabeled samples from diverse sources. We have collected 20,000+ images from real world (Material Recycling Facility, Open Food Facts) and online sources (web scraping, United States Plastic Corp.). This dataset serves as a valuable resource for enhancing the effectiveness of Machine Learning algorithms in plastic detection and identification. Our analyses incorporate various metrics, including Grad CAM, feature maps, and confusion matrices, providing insights into the specific information within an image that Machine Learning models leverage. In our analysis, we show that different aspects of plastic objects such as shape, label and background objects are being focused at, by the machine learning models. As a key takeaway, our findings suggest that integrating multiple modalities, such as spectroscopy in conjunction with computer vision, holds the potential to precisely discern plastic material characteristics, thereby further augmenting the accuracy of plastic sorting processes.
\section{Acknowledgements}\label{sec:acknowledgements}

We extend our sincere gratitude to our collaborators for their invaluable contribution to the formation of the plastic image dataset. Additionally, we would like to express our appreciation to the local Material Recovery Facility (MRF) for their collaboration, coordination, and generous permission that enabled us to collect images from the MRF conveyor belts. Their support and cooperation has been crucial to the progress of our research endeavors. This material is based upon work supported by the National Science Foundation (NSF) under Grant No. EFMA-2029375.

\bibliographystyle{ieeetr}
\setlength{\bibsep}{0pt}
\bibliography{plastics-sorting-visual}
\newpage
\section{About the Authors}\label{sec:about-the-authors}

\textbf{Vaishali Maheshkar} is a PhD student in the Department of Computer Science and Engineering at the University at Buffalo advised by Prof. Karthik Dantu. Her research interests are sensor systems, embedded systems, use of Machine Learning/AI to solve real world problems. (Presenter)
\\

\par\noindent \textbf{Aadarsh Anantha Ramakrishnan} is a third-year undergraduate student, studying Computer Science and Engineering at the National Institute of Technology, Tiruchirappalli. His research interests include Computer Vision, Deep Learning and Generative AI. Aadarsh is a PRACTICE-REU intern supported under the NSF-REU grant (EEC-2150424).
\\

\par\noindent \textbf{Charuvahan Adhivarahan} is a Postdoctoral Associate in the Department of Computer Science and Engineering at the University at Buffalo. His research interests are robotics perception, active mapping, exploration and in applying deep learning methods to solve real-world challenges in robotics.
\\

\par\noindent \textbf{Karthik Dantu} is an Associate Professor in Computer Science and Engineering at University at Buffalo. His research interests are in perception and coordination of mobile and multi-robot systems. At UB, he directs the Distributed RObotics and Networked Embedded Sensing (DRONES) Lab and is the Director of the University-Wide Center for Embodied Autonomy and Robotics. 
\end{document}